\documentclass[sigconf]{aamas} 
\usepackage{balance} % for balancing columns on the final page
\usepackage{fancyhdr,graphicx,amsmath}
\usepackage[ruled,vlined]{algorithm2e}
\usepackage{multirow}
\include{pythonlisting}

\newcommand{\algorithmicrequire}{ \textbf{Input:}}     %Use Input in the format of Algorithm
\newcommand{\algorithmicensure}{ \textbf{Output:}}    %UseOutput in the format of Algorithm

\setcopyright{ifaamas}
\acmConference[AAMAS '24]{Proc.\@ of the 23rd International Conference
on Autonomous Agents and Multiagent Systems (AAMAS 2024)}{May 6 -- 10, 2024}
{Auckland, New Zealand}{N.~Alechina, V.~Dignum, M.~Dastani, J.S.~Sichman (eds.)}
\copyrightyear{2024}
\acmYear{2024}
\acmDOI{}
\acmPrice{}
\acmISBN{}

\acmSubmissionID{367}

\title[AAMAS-2024 Formatting Instructions]{Solution-oriented Agent-based Models Generation with Verifier-assisted Iterative In-context Learning}

\author{Tong Niu}
\affiliation{
  \institution{Center for Brain-Inspired Computing Research, Tsinghua University}
  \city{Beijing}
  \country{China}}
\email{nt20@mails.tsinghua.edu.cn}

\author{Weihao Zhang}
\affiliation{
  \institution{Lynxi Technologies Co. Ltd}
  \city{Beijing}
  \country{China}}
\email{zwh18@mails.tsinghua.edu.cn}

\author{Rong Zhao}
\affiliation{
  \institution{Center for Brain-Inspired Computing Research, Tsinghua University}
  \city{Beijing}
  \country{China}}
\email{r_zhao@mail.tsinghua.edu.cn}

\begin{abstract}
Agent-based models (ABMs) stand as an essential paradigm for proposing and validating hypothetical solutions or policies aimed at addressing challenges posed by complex systems and achieving various objectives. This process demands labor-intensive endeavors and multidisciplinary expertise. Large language models (LLMs) encapsulating cross-domain knowledge and programming proficiency could potentially alleviate the difficulty of this process. However, LLMs excel in handling sequential information, making it challenging for analyzing the intricate interactions and nonlinear dynamics inherent in ABMs. Additionally, due to the lack of self-evaluation capability of LLMs, relying solely on LLMs is insufficient to effectively accomplish this process. In this paper, we present SAGE, a general solution-oriented ABM generation framework designed for automatic modeling and generating solutions for targeted problems. Unlike approaches reliant on expert handcrafting or resource-intensive neural network training, SAGE establishes a verifier-assisted iterative in-context learning process employing large language models (LLMs) to leverages their inherent cross-domain knowledge for tackling intricate demands from diverse domain scenarios. In SAGE, we introduce an semi-structured conceptual representation expliciting the intricate structures of ABMs and an objective representation to guide LLMs in modeling scenarios and proposing hypothetical solutions through in-context learning. To ensure the model executability and solution feasibility, SAGE devises a two-level verifier with chain-of-thought prompting tailored to the complex interactions and non-linear dynamics of ABMs, driving the iterative generation optimization. Moreover, we construct an evaluation dataset of solution-oriented ABMs from open sources. It contains practical models across various domains, completed with scenario descriptions and executable agent-based solutions. Evaluations by various LLMs demonstrate that SAGE leads to an average improvement of 18.7\% in modeling quality and 38.1\% in solution generation effectiveness. This work advances our understanding and ability in tackling complex real-world challenges across diverse domains through the application of ABM methodologies.
\end{abstract}


\newcommand{\BibTeX}{\rm B\kern-.05em{\sc i\kern-.025em b}\kern-.08em\TeX}

\begin{document}

%%% The following commands remove the headers in your paper. For final 
%%% papers, these will be inserted during the pagination process.

\pagestyle{fancy}
\fancyhead{}

%%% The next command prints the information defined in the preamble.

\maketitle 

%%%%%%%%%%%%%%%%%%%%%%%%%%%%%%%%%%%%%%%%%%%%%%%%%%%%%%%%%%%%%%%%%%%%%%%%

\section{Introduction}

Agent-based models (ABMs) provide a bottom-up perspective for exploring complex systems ~\cite{klugl2012agent}, such as societies ~\cite{macy2011social}, ecosystems~\cite{filatova2013spatial} and financial markets~\cite{turrell2016agent}. By modeling and simulating actions and interactions of autonomous agents within an artificial environment, ABMs can replicate real-world phenomena, validate hypotheses, uncover emergent behaviors, and formulate solutions or policies for specific problems. In the process of formulating solutions using ABMs, practitioners typically begin by modeling problem scenarios and subsequently simulating the hypothetical solutions proposed by experts to verify their effectiveness~\cite{watts2017should,lindblom1979usable}. This process poses significant challenges for practitioners ~\cite{xanthopoulou2019generating}, including the need for proficiency in abstract modeling to capture essential elements, structures and dynamics of the system to simplify real-world scenarios into an agent-based model. It also requires exceptional programming skills to translate the model into computational code. Most importantly, it demands a solid foundation of domain knowledge that entails a deep understanding of the problem context.

Leveraging advancements in computing methods~\cite{lecun2015deep,lazer2009computational}, substantial efforts are dedicated to automating the aforementioned modeling and solution generation process. Some initiatives apply intelligent algorithms, such as reinforcement learning~\cite{silver2016mastering}, to automatically generate strategies~\cite{zheng2022ai, koster2022human}. However, these approaches demand a prolonged learning processes reliant on explicit reward design or vast amounts of data. Some works make innovations by proposing domain-specific languages~\cite{tisue2004netlogo} or abstracting modeling principles~\cite{masad2015mesa} to simplify the programming complexity of models. Recently, large language models (LLMs) have shown significant progress in various tasks~\cite{kasneci2023chatgpt,thirunavukarasu2023large}, such as code generation~\cite{poesia2022synchromesh} and question answering ~\cite{daull2023complex}. Through extensive pre-training on diverse and abundant data, LLMs acquire the ability to generalize knowledge and apply it across various domains, positioning them as potentially valuable tools for the automated generation of solution-oriented ABMs. However, since LLMs are primarily designed to model natural languages, they excel in processing sequential information and linear causality~\cite{floridi2020gpt}. When applied directly to generate, analyze, and enhance ABMs, LLMs may encounter limitations due to their inherent design constraints in dealing with non-linear dynamics and intricate relationships among heterogeneous agents and the environment of ABMs. Besides, LLMs lack the self-checking and self-correction abilities for generated content. Therefore, ABMs generated by LLMs may contain factual inaccuracies, logical inconsistencies, and oversimplified intricacies of the real-world systems they aim to represent. Balancing the desire to streamline the solving process using LLMs with the need to ensure the executability and efficacy of the solution-oriented ABMs poses a formidable challenge.

%language patterns and semantics

To address these challenges, we present SAGE (Solution-oriented Agent-based models GEneration), a framework designed to leverage the in-context learning ability of LLMs for generating executable and verifiable agent-based solutions. SAGE comprises two stages: \textbf{"Modeling" stage} that uses a conceptual representation coupled with few-shot prompting techniques~\cite{qiao2022reasoning} to help LLMs comprehend the holistic structure and dynamics of the problem, ultimately facilitating the generation of executable ABMs that replicate the given problem scenario. \textbf{"Solving" stage} that employs iterative chain-of-thought (CoT)~\cite{wei2022chain} prompting with a two-level verifier that incorporates an objective representation to automatically generate and optimize effective solutions based on the above executable ABMs.

In SAGE, verifier-level1 identifies "compilation errors" and "lacking details" to ensure the executability and integrity of the ABM program. Meanwhile, verifier-level2 guide the iterative generation of CoT prompts, optimizing solutions by comparing objective criteria and simulation results. The CoT prompting breaks down the solution generation process into three steps: relations extraction, cause analysis, and solution proposal, enabling LLMs to formulate targeted solutions by comprehensively understanding the problem and the model. Both the conceptual representation (describing the problem scenario) and the objective representation (specifying the desired solution effects) serve as inputs to the LLM. These representations strike a balance between being semi-natural language (NL) for expression-diverse and user-friendly interface, and semi-structured for program accuracy and in-context learnability. By employing this approach, SAGE empowers LLMs to effectively organize intricate interactions within ABMs and leverage their inherent domain knowledge to propose efficient solutions. Furthermore, we have constructed an evaluation dataset comprising solution-oriented ABMs from open-source projects. Unlike existing code generation evaluation datasets, which primarily consist of partial code fragments accompanied by NL descriptions~\cite{wang2022execution,lu2021codexglue}, our dataset offers comprehensive scenario descriptions along with their ABM implementations. It encompasses real-world problem instances from various fields and their corresponding ABM-based solutions. The evaluation results of using SAGE with various state-of-the-art LLMs on this dataset demonstrate a massive improvement on generated ABMs' quality and problem-solving effectiveness, when compared to approaches that do not utilize the SAGE framework. The main contributions of our work are as follows:
 
 (1) SAGE: a framework that enables the automatic generation of executable and verifiable agent-based solutions through "Designing" and "Solving" stages. (2) The semi-structured conceptual and objective representations to formalize the ABM initialization and targets while maintaining the NL's expressivity. (3) A two-level verifier and CoT prompting method to streamline LLMs in generating effective solutions through iterative in-context learning. (4) A solution-oriented ABMs dataset to evaluate the quality of modeling and solution generation across multiple domains.

%%%%%%%%%%%%%%%%%%%%%%%%%%%%%%%%%%%%%%%%%%%%%%%%%%%%%%%%%%%%%%%%%%%%%%%%

\section{Related Work}

\subsection{Automatic Design and Generation of ABM}

The field of data and computational science has witnessed remarkable progress over the past decade, leading to a shift in the design and construction of ABMs from manual to automated approaches. This transition has empowered better exploration of data and expression of models. In terms of model structure generation, ~\cite{parsons2015automatic} proposes the MAGIC algorithm, which converts the raw observational data to a sequential compressed Markov decision process as agents' behavior and decision structures. Through a meta-programming system, ~\cite{xanthopoulou2019generating} can automatically generate executable ABM codes from an "Overview, Design concepts, and Details" (ODD) ~\cite{grimm2017documenting} description. ~\cite{greig2021generating} provides an evolvable approach to generate interpretable agent logic from scratch through program synthesis ~\cite{gulwani2017program} and genetic programming ~\cite{langdon2013foundations}, which endow users with transparency. Regarding the generation of agents' decisions, ~\cite{zheng2022ai} proposes a two level deep multi-agent reinforcement learning framework for automatic design of taxation policy. ~\cite{koster2022human} proposes a democratic AI capable of designing social mechanisms based on the majority preferences of humans through a human-in-loop reinforcement learning pipeline. These automatic generation methods often require extensive domain knowledge or long-term learning process based on massive data. Besides, some methods primarily focus on certain scenarios, lacking generality. Our approach overcomes these limitations with the ability of LLMs in efficiently handling multi-domain generation tasks, which is acquired during the large-scale pre-training phase. 

\subsection{LLMs for Code generation}

LLMs have achieved striking performance in a wide range of fields, including code generation tasks. Typically, these programming LLMs are developed by conducting self-supervised language modeling tasks on large unlabelled program language corpora followed by simple modifications and affordable fine-tuning. Specifically, three transformer architectures are widely developed: BERT ~\cite{devlin2018bert}, BART, and GPT ~\cite{radford2019language}. As for BERT, CodeBERT~\cite{feng2020codebert} is pre-trained on millions of program functions across six programming languages. GraphCodeBERT ~\cite{guo2020graphcodebert}, ContraBERT ~\cite{liu2023contrabert} are improved models that concern the inherent structure of code and enhance the robustness of CodeBERT. In terms of BART, PLBART ~\cite{ahmad2021unified} undergoes pre-training using a vast dataset comprising Java and Python functions along with their corresponding NL text, through denoising auto-encoding techniques. CommitBART ~\cite{liu2022commitbart} pre-trains BART using data collected from GitHub commits. Regarding GPT, CodeGPT ~\cite{lu2021codexglue} performs pre-training on Python and Java datasets sourced from the CodeSearchNet ~\cite{husain2019codesearchnet}. CodeX ~\cite{chen2021evaluating} conducts fine-tuning of GPT-3, which automatically fixed 28.8\% of programming problems in HumanEval datasets and is underlying technology behind GitHub Copilot ~\cite{Copilot}. These works have performed well on code snippet generation tasks. However, their efficacy tends to diminish in tasks like ABM programming that involve heterogeneous agents and intricate interaction patterns. To address this limitation, we augment LLMs' in-context learning by incorporating logically coherent and semi-structured representations.

\subsection{In-context Learning}

LLMs are capable of in-context learning (ICL) to learn tasks and generate desired output given only a few demonstration examples~\cite{dong2022survey}. ICL utilizes the analogy ability of the LLM and augments their task-specific working memory within the context, thereby influencing the likelihood of generating appropriate responses. Therefore, ICL can be applied during the inference procedure of a well-trained LLM, obviating the need for additional training. The organization of demonstrations within prompts, including their order and format significantly impacts ICL performance ~\cite{lu2021fantastically, zhao2021calibrate}. Many prompting methods have been proposed. ~\cite{visualprogramming} and ~\cite{zhou2022least} adopt manual construction such as providing template-based prompts and few-shot prompting. ~\cite{zhou2022large} and ~\cite{zhang2022automatic} leverage LLMs for automatic prompt generation. The CoT approach introduces intermediate reasoning steps to guide LLMs in mastering complex reasoning\cite{wei2022chain}. CoT has been widely used and developed since it was proposed~\cite{qiao2022reasoning}. Some researchers employ multi-stage ICL for CoT prompting and design CoT demonstrations for each step~\cite{zhou2022least, wang2022iteratively}. ~\cite{madaan2023self} and ~\cite{skreta2023errors} introduce an optimizer or verifier to calibrate the generation result through iterative generation and optimization procedures. ICL has found applications in various domains, including visual tasks ~\cite{visualprogramming}, language reasoning ~\cite{qiao2022reasoning}, and task planning ~\cite{skreta2023errors}. However, there is limited research on utilizing ICL to guide LLMs to generate high quality ABMs. 

%%%%%%%%%%%%%%%%%%%%%%%%%%%%%%%%%%%%%%%%%%%%%%%%%%%%%%%%%%%%%%%%%%%%%%%%

\section{SAGE}

\begin{figure}[h]
  \centering
  \includegraphics[width=1\linewidth]{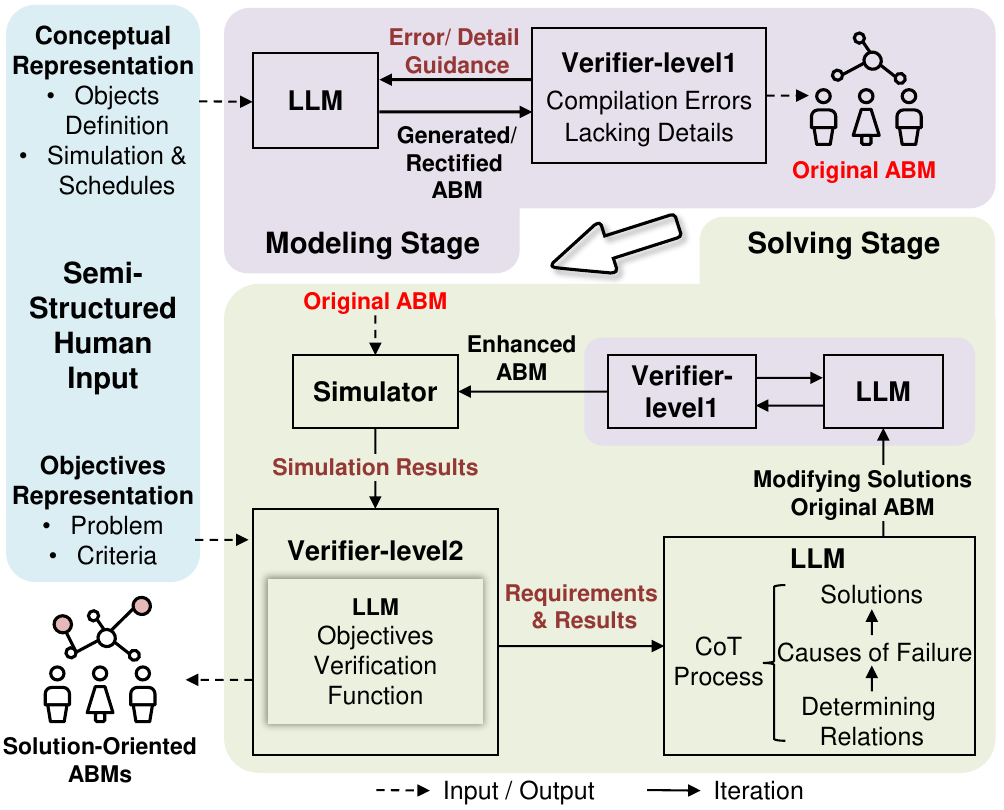}
  \caption{The workflow of SAGE.}
  \label{fig:workflow}
  \Description{}
\end{figure}

\textbf{Architecture and Workflow} SAGE divides the solution generation process into two stages, as shown in Figure~\ref{fig:workflow}. The initial stage, termed the \textbf{"Modeling" stage}, aims to generate an executable ABM to simulate a given problem scenario. During this stage, the LLM creates the ABM program based on the conceptual representation of the scenario provided by users. This semi-structured representation explicitly depicts the intricate interactions and dynamic events within ABMs through its structured components, facilitating analysis by LLMs. Besides, it enables users to describe scenarios using more expressive natural language. The generated ABM then undergoes automatic verification for both executability and integrity by the verifier-level1. The LLM iteratively rectifies the program based on the verification results until a correct ABM is obtained. Such verifier-assisted iteration compensates for the self-checking deficiencies in LLMs, avoiding execution errors and oversimplifications.

% 忽然出现的"compilation error" and "lacking details"让人感觉不是很连贯，建议后面再说这些细节。但lli里有啊
% Therefore, the "compilation error" and "lacking details" detection capabilities of verifier-level1 compensate for the self-checking deficiencies in LLMs.

% facilitating analysis and further expansion by LLMs.

Subsequently, the process transitions to the \textbf{"Solving" stage} that focuses on generating effective solutions for given problems. In this stage, users specify the target problem and desired effects or constraints of the solution using objective representations. (Both conceptual and objective representations are concatenated with examples to form input prompts for ICL.) Then, verifier-level2 is designed to utilize LLMs to generate verification functions based on the objective representations. These functions are used to verify whether simulation results meet the specific objectives, which drives the optimization of solutions and compensates for the self-evaluation deficiencies in LLMs. Based on the verification results, verifier-level2 provides CoT instructions that guide LLMs in generating solutions step by step, which include relations extraction, cause analysis, and solution proposal. This CoT design guides LLMs in recognizing the holistic structure of the current ABM and subsequently formulating more specific solutions based on relevant elements, thereby mitigating the limitations of LLMs in deducing the complex interaction dynamics of ABMs solely from static simulation results. The generated solutions are integrated back into the existing ABM and undergo further iterative rectification using verifier-level1 until the ABM with new solutions is executable and comprehensive. This process continues iteratively between verifier-level1 and verifier-level2 until the desired objectives are met. It is worth noting that these two stages can be decoupled and independently utilized depending on research purposes, allowing flexibility in applications. The pseudocode outlining the "Modeling" stage and "Solving" stage can be found in Algorithm 1 and Algorithm 2.

% This ABM will be employed to simulate the actual effects of the proposed solution within the defined problem scenario. 

\begin{figure*}[t!]
  \centering
  \includegraphics[width=1\linewidth]{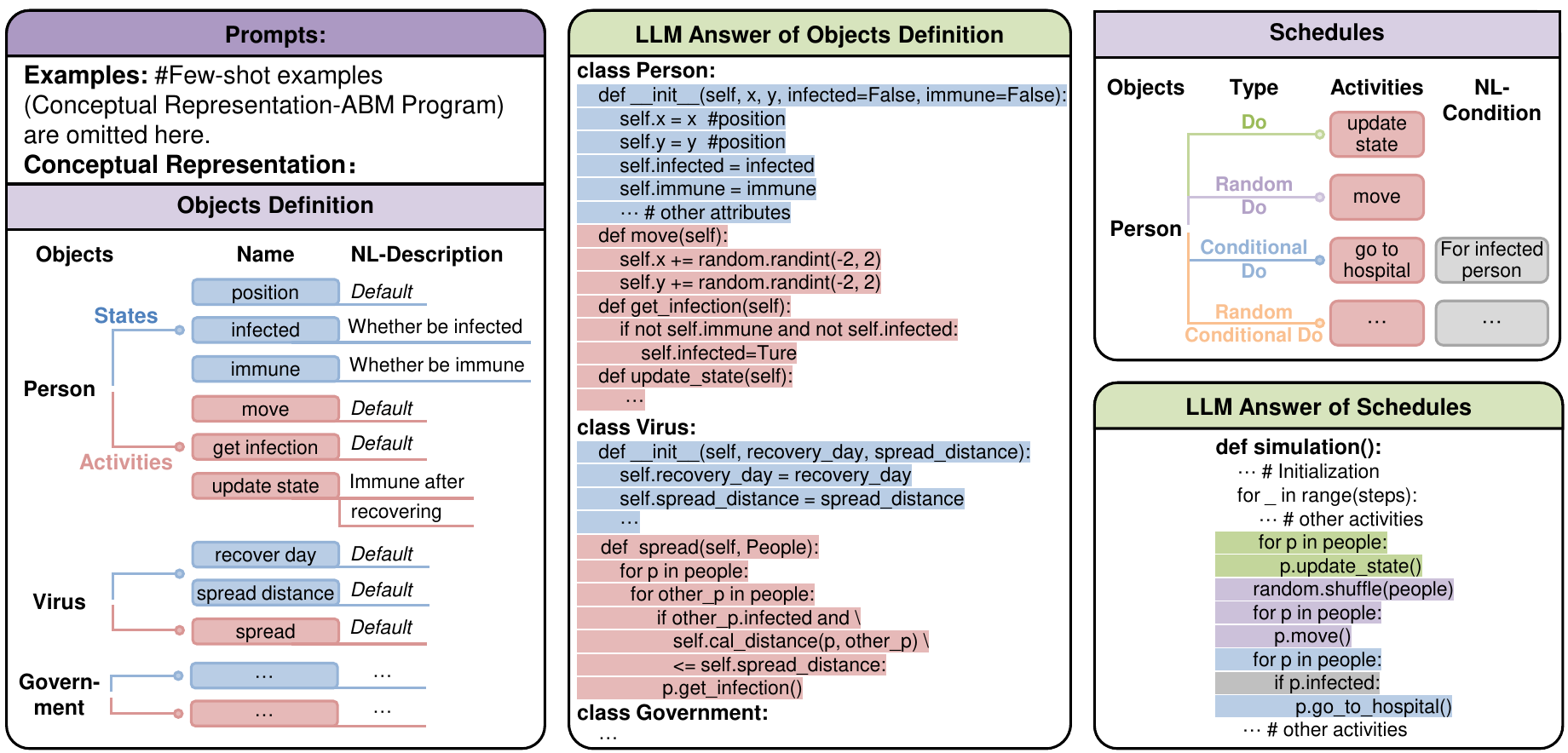}
  \caption{Conceptual representation for ABM description and the generated ABM program.}
  \label{fig:conceptual_ABM}
  \Description{}
\end{figure*}

\subsection{"Modeling" Stage}

\textbf{ABM Conceptual Representation.} ABM involves intricate elements such as agents, environments, and their interactions. These elements typically can be instantiated as object states, activities comprising conditions and algorithms for state updating, as well as the scheduling of these activities. However, translating these concepts into practical ABMs requires significant efforts and domain expertise, often proving to be abstract and daunting for non-programmers. To bridge this gap, we design a semi-structured, semi-natural language conceptual representation that serves as a user-friendly guide, allowing users to provide detailed scenario descriptions essential for developing ABM models. Meanwhile, leveraging the power of few-shot prompting, LLMs can learn to transform and expand the information encapsulated in the conceptual representation into executable ABM programs.

Figure~\ref{fig:conceptual_ABM} illustrates a conceptual representation and a generated ABM program. According to the above ABM necessities, the representation consists of two structured components. The first component, "object definition", corresponds to the states and activities of agents and environment objects. The second component, known as "scheduling", corresponds to the execution sequence or trigger conditions of activities. In the object definition, "states" define the descriptive attributes of an object at a specific time step, while "activities" define the dynamic interactions involving the object. For instance, when defining a person in the scenario of epidemic transmission, the states may include "age", "infected", and "immune", while the activities may include "get\_infected", "get\_immune" and so on. The specific meanings of each state and activity are presented by natural language (NL). Guided by the few-shot prompt (shown in supplementary material), the LLM formalizes the NL descriptions and generalizes the omitted descriptions by utilizing its inherent knowledge and contextual information. In the current implementation, conceptual representations are presented in JSON format. 

The simulation process of an ABM involves the flexible scheduling of different agents, each executing their internal functions in a specific order and under certain conditions. To simplify the description of the scheduling of the aforementioned object definition, we abstract and structure them as schedule primitives: \textit{Do(Object\_name, Activity\_name)} means that all instances of this object perform this activity in the default order; \textit{Random\_Do(Object\_name, Activity\_name)} means that all instances of this object perform this activity in a random order; \textit{Conditional\_Do(Object\_name, Activity\_name, conditions)} means that instances satisfying the conditions of this object perform this activity in the default order; \textit{Random\_Conditional\_Do( Object\_name, Activity\_name, conditions)} means that all instances satisfying the conditions of this object perform this activity in a random order. Figure~\ref{fig:conceptual_ABM} shows an example illustrating the corresponding ABM program with schedule primitives. 

\begin{algorithm}[t]
\SetAlgoLined
\algorithmicrequire{Conceptual representation of scenario $\mathcal L$, max number of iterations $i$ }\;
\algorithmicensure{ABM program $A$}\;
 $P_A$ = Prompt\_gen\_abm($\mathcal L$)\;
 $A$ = LLM\_generator($P_A$)\;
 defects = Verifier\_level1($A$)\;
 \While{len(defects)>0 and $i$>0}{
   $P_D$ = Prompt\_defects($A$, defects)\;
   $A$ = LLM\_generator($P_D$)\;
   defects = Verifier\_level1($A$)\;
   $i$ = $i$-1
 }
 \caption{Scenario-specific ABM generation in "Modeling" stage}
\end{algorithm}

\begin{figure*}[t!]
  \centering
  \includegraphics[width=1\linewidth]{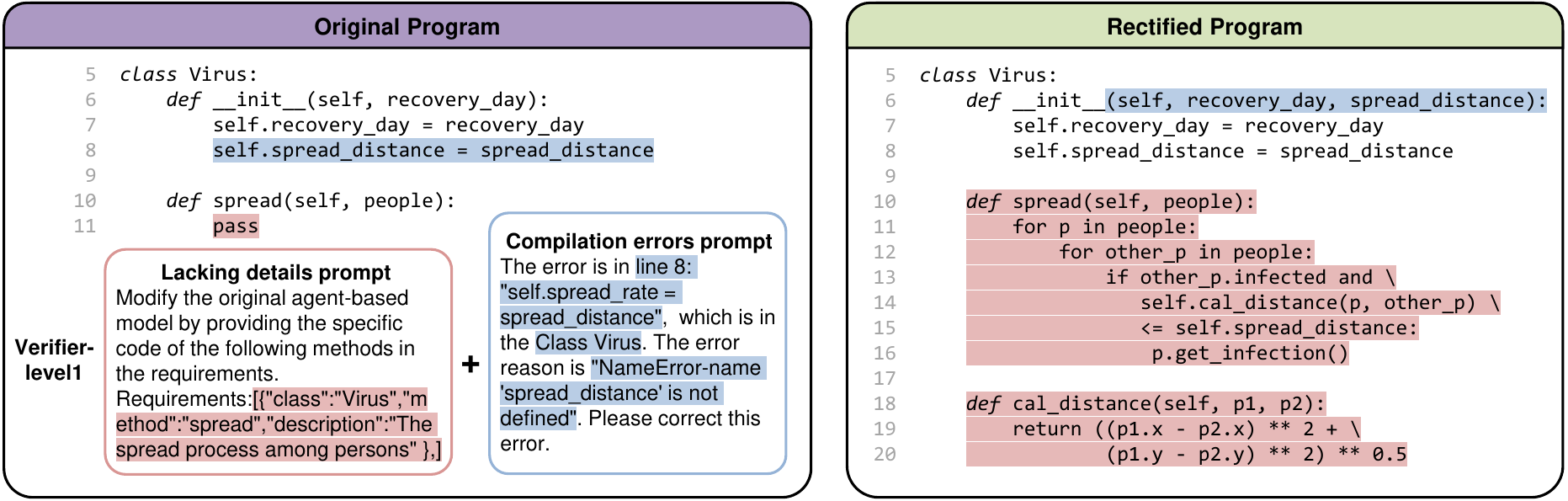}
  \caption{The rectification prompts generated by verifier-level1 and the rectified ABM program.}
  \label{fig:verifier1}
  \Description{}
\end{figure*}

\textbf{Verifier-level1: Program correctness verification.} In light of the challenges posed by the lack of self-checking capabilities in LLMs, we devise an automatic ABM program verifier to identify defects in generated programs, ensuring their executability and integrity. These defects are then filled in a prompt template for LLM to further rectify the ABM program, enabling iterative rectification until an executable ABM with complete details is obtained. Defects are classified into two types: "compilation errors" and "lacking details". "Compilation errors" hinder proper model execution. For Python programs, they indicate interpretation errors and certain runtime errors. To rectify these errors, verifier-level1 invokes the default compiler or interpreter to compile or run the ABM program, analyzing the returned error messages to assemble the "compilation errors" rectification prompts for LLMs. Each error message is structured in the prompt as $[error\_line, error\_code, error\_reasons]$. On the other hand, defects falling under "lacking details" refer to empty classes or functions that lack substantive implementation details. These defects are often caused by using placeholders instead of detailed code, leaving specific interaction behaviors and updated variable values unaddressed. Verifier-level1 tackles these issues systematically. First, it identifies these defects through text analysis, linking activity names to object classes. This information is then merged with the corresponding activity descriptions from the aforementioned conceptual representation, providing structured guidance for prompt-based rectification. An example of the verifying and rectifying process is shown in Figure~\ref{fig:verifier1}. 

\subsection{"Solving" Stage}

\textbf{Objective representation.} Under the specific ABM generated by the "Modeling" stage, users may seek solutions or formulate policies addressing distinct problems, such as decreasing the spread rate of a virus in an epidemic model. We design an objective representation, which is also a semi-structured and semi-NL representation, to effectively convey these requirements. The representation comprises two parts: the description of the target problem (e.g. "decrease the spread rate of the virus") and the criteria for judgment. The judgment criteria encompass not only the intended outcomes of the solutions (e.g. "The spread rate of virus is reduced to below 0.1") but also any restrictions (e.g. "The ground truth, like spread distance, should not change"). The target problem is expressed in NL, providing a clear understanding of the specific issue at hand. On the other hand, the criteria typically involve variables that can be quantitatively or qualitatively compared within the described problem scenario. Each criterion is structured as $[variable_{name}, variable_{example}, requirement]$, where $variable_{example}$ specifies the data type of the variable, and $requirement$ described in NL. 

\begin{figure}[h]
  \centering
  \includegraphics[width=1\linewidth]{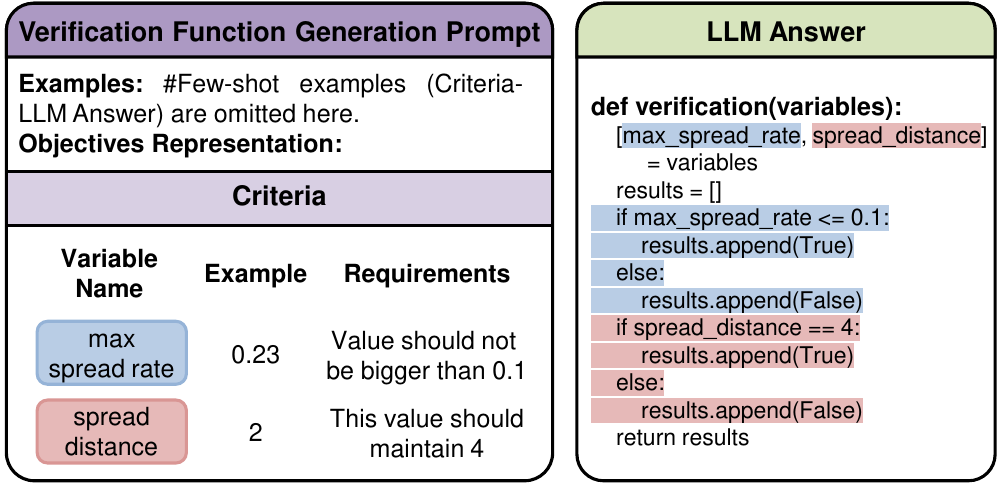}
  \caption{Objective representation for desired solution effects and the generated verification program.}
  \label{fig:verifier2}
  \Description{}
\end{figure}

\begin{algorithm}[h]
\SetAlgoLined
\algorithmicrequire{objective representation $\mathcal O$, original executable ABM $A$, max number of iterations $i$ }\;
\algorithmicensure{ ABM program $A$, solutions}\;
 $\mathcal R$ = Simulator($A$) \# real results \; 
 $P_V$ = Prompt\_gen\_Veri($\mathcal O$)\;
 Veri\_func = LLM\_generator($P_V$)\;
 satisfying\_flag = Verifier\_level2(Veri\_func, $\mathcal R$)\;
 \While{satisfying\_flag==False and $i$>0}{
   $P_C$ = Prompt\_CoT($\mathcal R$, $\mathcal O$, $A$)\;
   relations, reasons, solutions = LLM\_generator($P_C$)\;
   $P_M$ = Prompt\_modification($A$, solutions)\;
   $A$ = LLM\_generator($P_M$)\;
   $i$ = $i$-1\;
   defects = Verifier\_level1($A$)\;
    \While{len(defects) > 0}{
        $P_D$ = Prompt\_defects($A$, defects)\;
        $A$ = LLM\_generator($P_D$)\;
        defects = Verifier\_level1($A$)\;
    }
    $\mathcal R$ = Simulator($A$) \# real results \; 
   satisfying\_flag = Verifier\_level2(Veri\_func, $\mathcal R$)\;
   \If{$i$ == 0}{
      break\;
   }
 }
 \caption{Scenario-specific solution generation in "Solving" stage}
\end{algorithm}

\begin{figure*}[h]
  \centering
  \includegraphics[width=1\linewidth]{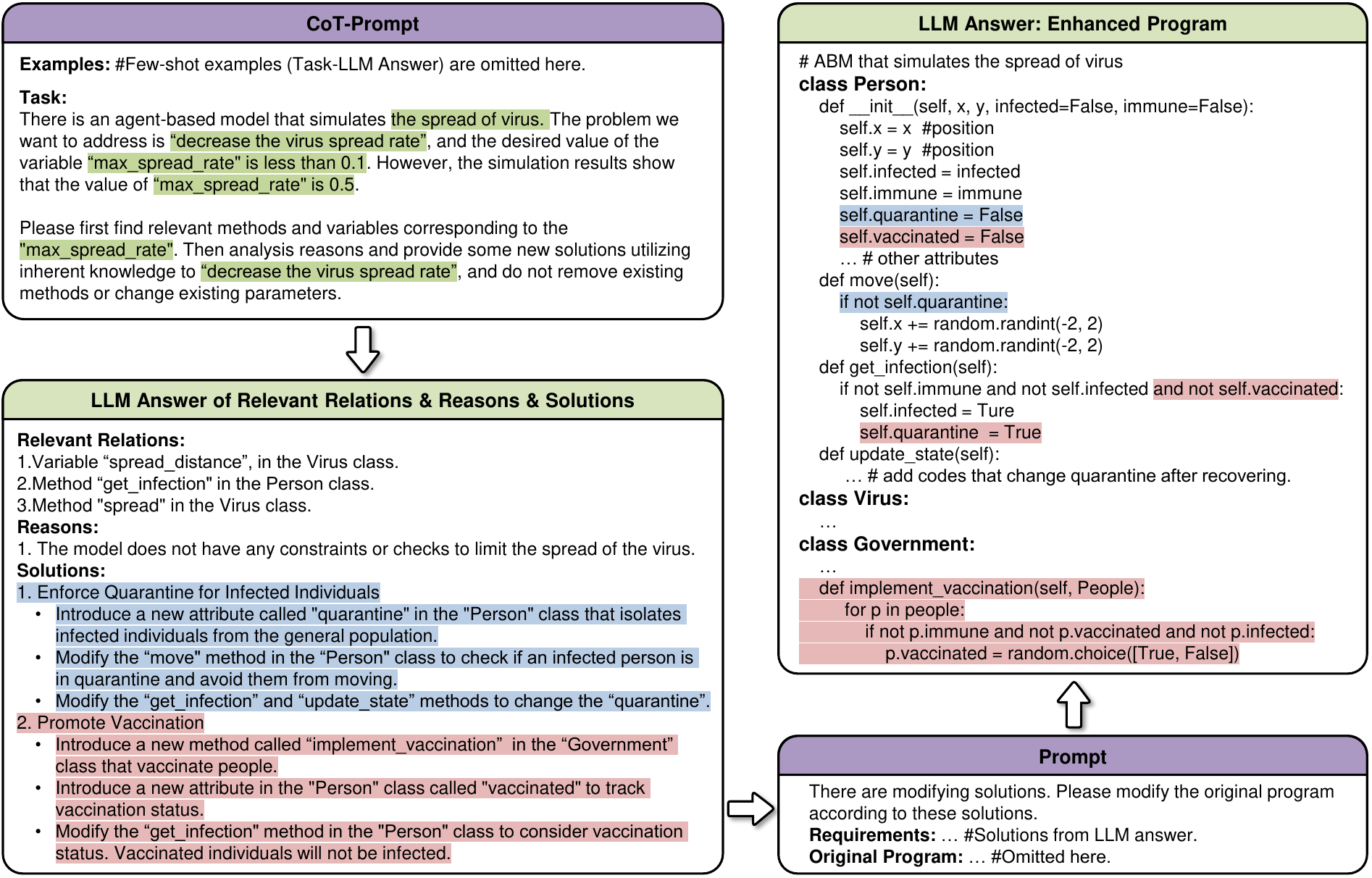}
  \caption{The CoT workflow and the generated solutions with enhanced ABM program.}
  \label{fig:cot}
  \Description{}
\end{figure*}

\textbf{Verifier-level2: Solution effectiveness verification.} Verifier-level2 is designed to assess the efficacy of solutions. Using a few-shot prompt, the LLM converts the problem and criteria in the objective representation into verification functions. These functions are crucial as they assess whether the variables extracted from simulation results meet the verification criteria. If all criteria are satisfied, verifier-level2 acknowledges these solutions as successfully verified. Consequently, the executable ABM embedded with these solutions is regarded as the final output of SAGE. Figure~\ref{fig:verifier2} showcases a criteria prompt and the corresponding verification function answer for the problem "decrease the spread rate of the virus". Please note that the template of the verification function, including the format of input and output, is provided in the examples within the prompt (omitted in Figure ~\ref{fig:verifier2}). This semi-structured representation combines the flexibility of natural language for describing criteria with the clarity provided by a list structure, allowing for the precise expressions of different criteria. These expressions can be any attribute or complex function (including functions over time series), provided that you record them during simulation and can return the records to the verification function after simulation completion. Moreover, the verification functions generated by the LLM are written in general-purpose programming languages, enhancing their versatility and comprehensiveness in comparison to combinatorial arrangements of pre-defined criteria implementation. 

\textbf{Solution generation with CoT prompt.} 
Verifier-level2 utilizes the simulation results that fail to meet the evaluation criteria to generate a prompt for the next-stage LLM to enhance the ABM program. Because ABMs involve non-linear dynamics and complex spatiotemporal interactions, it is challenging for LLMs to determine how to modify a model to meet the criteria based solely on a static experimental result. The CoT prompt guides the model through a series of logically connected thinking steps, enabling coherent reasoning and logical thinking. Therefore, we apply the CoT prompt technique to enhance the solution generation of LLMs. Given that the individual state changes in an ABM are fundamentally driven by interactions with the current states of other individuals, our CoT is designed with three steps: relations extraction, which guides LLMs in extracting the operations within the ABM program that influence the evaluated variables; cause analysis, which guides LLMs in analyzing the extracted operations to identify the reasons for unsatisfied criteria; and solution proposal, which guides LLM in leveraging domain knowledge to propose solutions based on the identified causes. Figure ~\ref{fig:cot} illustrates an example on seeking solutions of "decrease spread rate of virus". Following the guidance of the CoT, the LLM first extracts the relevant functions and variables that impact the spread rate, such as variables "spread\_distance" and the function "spread". Then, the LLM provides an analysis of the reasons, e.g. lack of constraints or checks contributes to the uncontrolled spread of the virus. Lastly, the LLM proposes two solutions: "enforce quarantine" and "promote vaccination" accompanied by instructions on how to implement them within the original ABM.

%%%%%%%%%%%%%%%%%%%%%%%%%%%%%%%%%%%%%%%%%%%%%%%%%%%%%%%%%%%%%%%%%%%%%%%%
\section{Dataset and Evaluation}

\subsection{Solution-oriented ABMs Dataset}
While several code datasets are available for training or testing LLMs' programming abilities, the majority of them typically consist of isolated code snippets without sufficient contextual information or background knowledge. In particular, there is a lack of evaluation datasets tailored to the generation of ABMs for problem-solving assistance. In this work, we introduce a solution-oriented ABMs dataset aimed at evaluating the utility and effectiveness of the generated ABMs in problem-solving scenarios. This will also promote research and advancements in the fields of natural language processing and programming. 

The evaluation dataset is primarily divided into two sub-datasets. The first is the "scenario-model" dataset, which mainly evaluates the language model's ability to generate corresponding ABM programs based on the given scenario descriptions. It contains 50 samples, each consisting of a natural language-based scenario description, a conceptual representation-based scenario description as a task, and an executable ABM that programs the scenario from the task description as a ground truth answer. The second part is the "problem-solution" dataset, which primarily assesses the LLMs' capability to propose solutions based on the given ABM for specific problems. This sub-dataset comprises 30 samples, each consisting of a problem description and an executable original ABM as a task, and an executable ABM that comprises the proposed solutions as a ground truth solution answer. In this sub-dataset, some samples contain strict criteria (criteria-defined samples), while others without criteria are open-ended samples. The dataset spans various domains as shown in Table ~\ref{tab:dataset} and is primarily sourced from open-source projects on GitHub. Descriptions of the scenarios and problems are derived from project introductions and comments within the code, which are subsequently supplemented and standardized through manual efforts. The code segments in the dataset are extracted from the code of corresponding projects, meticulously organized and refined. 

\begin{table}[]
\caption{Dataset description}
	\label{tab:dataset}
\begin{tabular}{c|c}
\hline
\multirow{2}{*}{\begin{tabular}[c]{@{}c@{}}\\scenario-model \\ dataset\end{tabular}} & Domains \\ \cline{2-2} 
 & \begin{tabular}[c]{@{}c@{}}Public health: 22\%; Society: 22\%;\\ Ecology: 12\%; Traffic: 10\%; \\ Economics: 10\%; Biology: 10\%; \\ Education: 8\%; Gaming: 6\%;\end{tabular} \\ \hline
\begin{tabular}[c]{@{}c@{}}problem-solution\\ dataset\end{tabular} & \begin{tabular}[c]{@{}c@{}}Criteria-defined samples: 40\%\\ Open-ended samples: 60\%\end{tabular} \\ \hline
\end{tabular}
\end{table}

\begin{table*}
	\caption{Evaluation results on executable ABMs generation}
	\label{tab:model_evaluation}
\begin{tabular}{ccccccc}
\hline
LLMs         & \begin{tabular}[c]{@{}c@{}}CodeBLEU\\ No-SAGE\end{tabular} & \begin{tabular}[c]{@{}c@{}}CodeBLEU\\ SAGE\end{tabular} & \begin{tabular}[c]{@{}c@{}}Executable\\ No-SAGE\end{tabular} & \begin{tabular}[c]{@{}c@{}}Executable\\ SAGE\end{tabular} & \multicolumn{1}{c}{\begin{tabular}[c]{@{}c@{}}Elaborate\\ No-SAGE\end{tabular}} & \multicolumn{1}{c}{\begin{tabular}[c]{@{}c@{}}Elaborate\\ SAGE\end{tabular}} \\ \hline
GPT-3.5-Turbo  & 42.47 & 52.18 & 64.00 & 90.00 & 68.00 & 90.00\\
Code-Llama-34b & 44.94 & 53.36 & 64.00 & 86.00 & 78.00& 88.00\\
Claude-instant & 42.49 & 49.64 & 48.00 & 82.00 & 74.00 & 84.00\\
GPT-4          & 45.73      & 53.29      & 78.00      & 96.00      & 74.00 & 96.00 \\ \hline
\end{tabular}
\end{table*}

\subsection{"Modeling" Evaluation}
To evaluate the practical applicability of SAGE in modeling, we conduct experiments using the "scenario-model" sub-dataset. Four state-of-the-art LLMs: GPT-3.5-Turbo, Code-Llama-34b, Claude-instant, and GPT-4 are chosen as the generative models. These LLMs generate ABMs according to different scenario descriptions from the dataset under two conditions: one involving the generation process with SAGE, and the other without SAGE (No-SAGE). Specifically, no-SAGE process refers to LLMs generating ABMs and solutions solely based on natural language descriptions without using SAGE's semi-structured representation or two-level verification optimization. The generated ABMs are then compared against their corresponding ground truth models to evaluate the generation quality. To ensure fairness, given SAGE's iterative nature in the generation process, each sample undergoes 10 times of modeling in the No-SAGE condition. The best of these results is then used for subsequent comparisons. The "10" is same with the maximum number of iterations set in the SAGE-involved experiments.

\begin{figure}[t]
  \centering
  \includegraphics[width=0.9\linewidth]{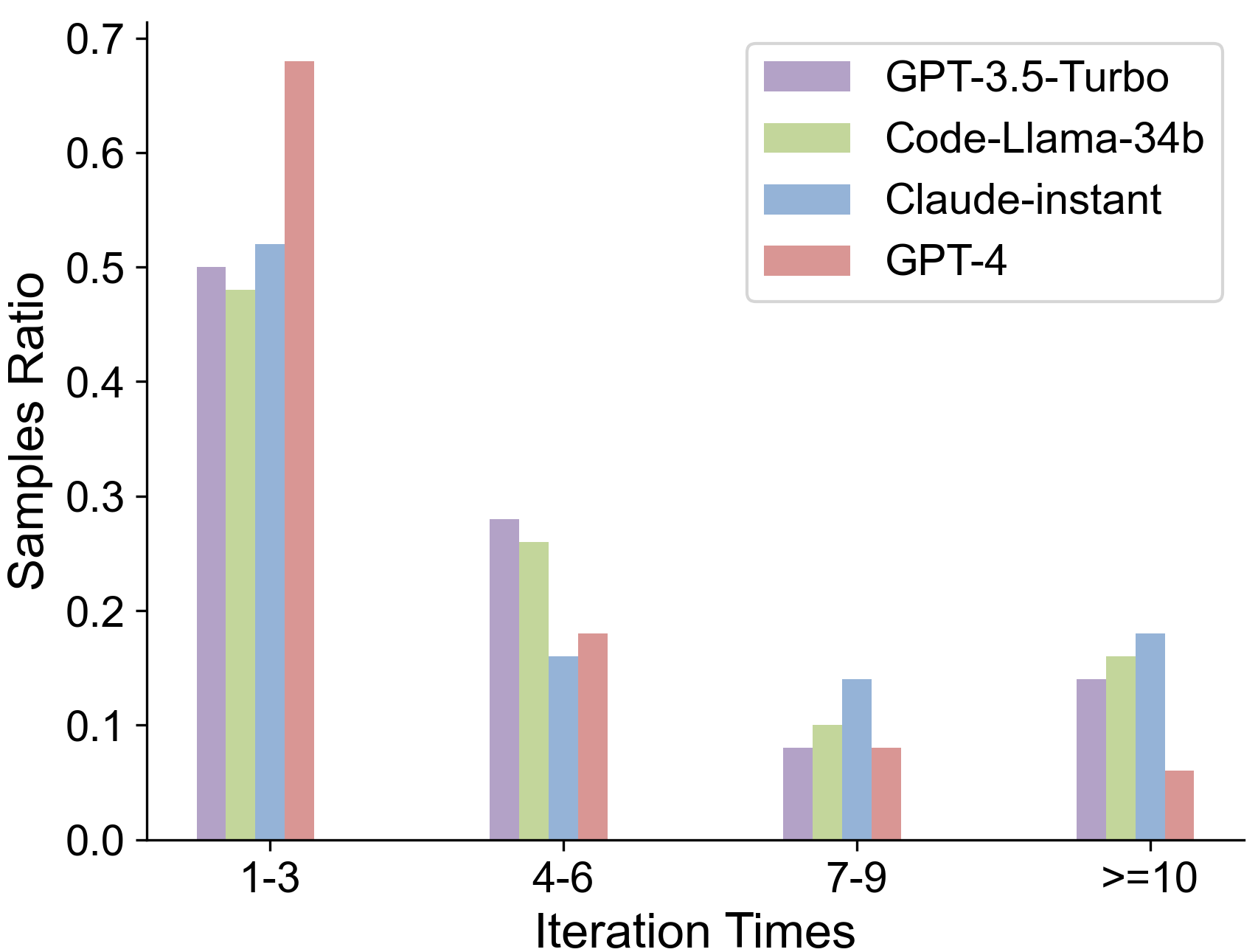}
  \caption{Iterations times for ideal ABMs generation}
  \label{fig:exp1}
  \Description{Iterations times for executable ABMs generation, there are four classifications, less than three times, 4-6 times, 7-9 times and more than ten times.}
\end{figure}

\textbf{Evaluation metrics} Two metrics are employed to evaluate the quality of the generated ABMs. The first metric involves measuring the similarity between the generated model and the ground truth answer using CodeBLEU\cite{chen2021evaluating}, which weights the average of the lexical, abstract syntax tree (AST), and data flow match between the generated code and the ground truth. We reduce the weight given to lexical match and improve the weight given to AST and dataflow match during the evaluation, since the sequence of object definitions does not impact the ultimate simulation outcomes. The second metric evaluates whether the generated ABMs are executable and elaborate, which is conducted by the verifier-level1 of SAGE.

\textbf{Evaluation results} 
The first two columns of Table ~\ref{tab:model_evaluation} indicate that ABMs produced using SAGE with LLMs consistently yield higher scores (four LLMs' average improvement of 18.7\%) compared to those generated by No-SAGE, when applying CodeBLEU for assessment.From the perspective of the executability and integrity of the generated ABMs, the last four columns of Table ~\ref{tab:model_evaluation} present the ratios of the generated ABMs that exhibit these qualities among all answers. These data underscore SAGE's remarkable efficacy in facilitating error correction and content enrichment of ABMs, as evident across all four LLMs. Figure ~\ref{fig:exp1} illustrates the number of iterations required for different LLMs to generate an executable and elaborate ABM when utilizing SAGE (Generation failed samples are also included in the category of 10 or more iterations). The results show that with the incorporation of SAGE, over 70\% of the samples can be resolved within six iterations or fewer. 

\subsection{"Solving" Evaluation}

In the "Solving" evaluation experiments, we employ the "problem-solution" dataset as the testing set and utilize two general-purpose LLMs, GPT-3.5-Turbo and GPT-4, as the solution generation models. The impact of SAGE on the successful resolution of the problem is assessed by comparing the success rates of problem resolution between the scenarios with and without the use of SAGE. Similar to the "modeling" evaluation experiments, each sample undergoes 5 times of "solving process" in the No-SAGE condition, and the best solution-oriented ABM is chosen. Besides, to assess the relative importance of different verification levels in the "solving" process, we analyze the failure reasons in all unsuccessful cases.

\textbf{Evaluation metrics} In addition to assessing the executability of the generated solutions, for samples with pre-defined criteria, we determine the success of problem resolution based on whether the simulation results meet the evaluation criteria. For samples without specific criteria, which are named as open-ended problems, we manually evaluate solutions' practical significance, which primarily consist of two standards. Firstly, from the perspective of object definitions, we examine whether there are meaningful additions or deletions of states and activities of objects, rather than simple modifications to the parameters involved in the existing ABMs. Secondly, from the perspective of scheduling, we assess whether the added or deleted states or activities truly affect the simulation process, in other words, whether they are invoked during the execution process. The solutions of one sample is confirmed only when all of these standards are simultaneously fulfilled.

\begin{table}[h]
\caption{Problem solving-rates on Criteria-defined(CD) samples and Open-ended(OE) samples}
	\label{tab:SER}
\begin{tabular}{ccccc}
\hline
\begin{tabular}[c]{@{}c@{}}LLMs\end{tabular} &
  \begin{tabular}[c]{@{}c@{}} CD\\ No-SAGE\end{tabular} &
  \begin{tabular}[c]{@{}c@{}} CD\\ SAGE\end{tabular} &
  \begin{tabular}[c]{@{}c@{}} OE\\ No-SAGE\end{tabular} &
  \begin{tabular}[c]{@{}c@{}} OE\\ SAGE\end{tabular} \\ \hline
GPT-3.5-Turbo &   58.33&  91.67&  38.89&  88.89\\
GPT-4 &  41.67&  100&  66.67&  77.78\\ \hline
\end{tabular}
\end{table}

\begin{figure}[h]
  \centering
  \includegraphics[width=1\linewidth]{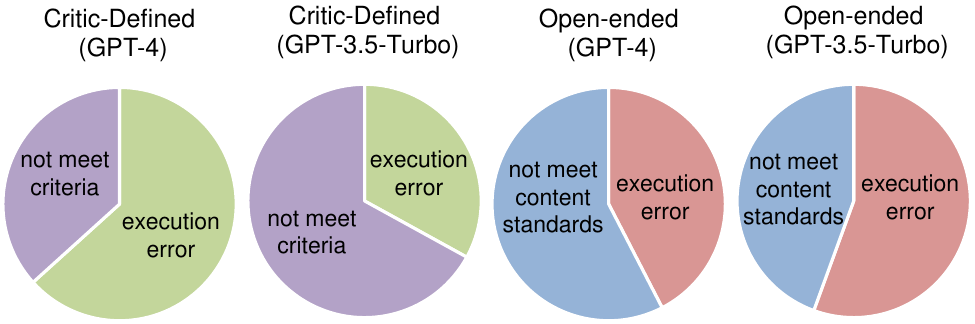}
  \caption{Causes statistics for unsuccessful solutions}
  \label{fig:exp2}
  \Description{}
\end{figure}

\textbf{Evaluation results} The evaluation results (Table ~\ref{tab:SER}) indicate the problem-solving rates. It is evident that when comparing problem solving rates with and without SAGE involvement, there is a noticeable enhancement in problem-solving efficiency under the SAGE involvement for both criteria-defined samples and open-ended samples (average improvement of 38.1\% ). The verifier-level1 proves effective in preventing execution errors, while the verifier-level2 facilitates the generated solutions meet the pre-defined criteria and content standards. Figure ~\ref{fig:exp2} shows that both types of above defects cannot be overlooked. Therefore, in order to guarantee the generation of executable and effective solutions, both levels of verifiers are indispensable.

%%%%%%%%%%%%%%%%%%%%%%%%%%%%%%%%%%%%%%%%%%%%%%%%%%%%%%%%%%%%%%%%%%%%%%%%
\section{Conclusions}

In this paper, we propose a solution-oriented ABM generation framework: SAGE. SAGE utilizes the powerful abilities of LLMs, including in-context learning, natural language understanding, compositional generalization, and domain-knowledge retrieval. This enables the development of a two-stage, iterative procedure for automatic agent-based modeling and solution generation. Our results demonstrate a significant improvement in generation performance compared to traditional No-SAGE approaches. Notably, SAGE achieves this without requiring the extensive training process of LLM and a large amount of data. Thus, SAGE has the ability to significantly reduce the barriers to creating ABMs and utilizing them for problem-solving, enhancing the productivity of complex system research.

\begin{acks}
This work was partly supported by the National Key Research and Development Program of China (grant no. 2021ZD0200300), National Nature Science Foundation of China (nos. 62088102 and 61836004), IDG/McGovern Institute for Brain Research at Tsinghua University
\end{acks}

%%%%%%%%%%%%%%%%%%%%%%%%%%%%%%%%%%%%%%%%%%%%%%%%%%%%%%%%%%%%%%%%%%%%%%%%

%%% The next two lines define, first, the bibliography style to be 
%%% applied, and, second, the bibliography file to be used.

%\bibliographystyle{ACM-Reference-Format} 
%\bibliography{sample}

%%%%%%%%%%%%%%%%%%%%%%%%%%%%%%%%%%%%%%%%%%%%%%%%%%%%%%%%%%%%%%%%%%%%%%%%

%%% -*-BibTeX-*-
%%% Do NOT edit. File created by BibTeX with style
%%% ACM-Reference-Format-Journals [18-Jan-2012].

\end{document}